\documentclass[letterpaper, 10 pt, conference]{ieeeconf} %

\IEEEoverridecommandlockouts                              %

\usepackage{amsmath} %
\usepackage{amssymb}  %
\usepackage{amsfonts}       %
\usepackage{bm}
\usepackage{cite}
\usepackage{url}
\usepackage{siunitx} %
\usepackage{subcaption}
\usepackage{graphicx}
\usepackage{footnote}

\usepackage{color,soul}

\usepackage{balance}
\usepackage{booktabs}
\usepackage{multirow}
\usepackage{xcolor}
\usepackage[normalem]{ulem}   %

\newcommand{\tp}{\mathsf{T}}
\NewDocumentCommand{\xtheta}{o}{{\bm{x}^{\theta}\IfValueT{#1}{_{#1}}}}
\newcommand{\bx}{\bm{x}}
\newcommand{\bs}{\bm{s}}
\newcommand{\ba}{\bm{a}}
\newcommand{\by}{\bm{y}}
\newcommand{\bw}{\bm{w}}

\title{\LARGE \bf
  Pose Estimation for Peg Insertion from Proprioceptive and Haptic
  Feedback via Learned Contact-Conditional Measurement Models
}
\author{Zongyao Yi\authorrefmark{1}, Joachim
  Hertzberg\authorrefmark{2}\authorrefmark{1} and
  Martin Atzmueller\authorrefmark{2}\authorrefmark{1}%
  \thanks{\authorrefmark{1}German Research Center for Artificial
    Intelligence, DFKI Niedersachsen, Osnabrück, Germany
    {\tt\small zongyao.yi@dfki.de}
  }%
  \thanks{
    \authorrefmark{2}Osnabrück University, Institute of Computer
    Science, Osnabrück, Germany
  }%
}

\begin{document}
\bstctlcite{BSTcontrol}

\maketitle
\thispagestyle{empty}
\pagestyle{empty}

\IEEEpeerreviewmaketitle

\begin{abstract}
  Prior work~\cite{yi2025learning} presented a learnable
  physics-based predictive model of contact dynamics, and
  demonstrated its use as a dynamics model for model-predictive
  control in peg insertion, assuming prior knowledge of the socket
  pose. This article extends that method with contact-conditional
  probabilistic prediction and investigates its role as a measurement
  model. The extended model
  predicts end-effector motion and wrench readings together with
  their full covariance conditioned on the contact state. These predictions
  are then used to calculate the likelihoods of the measurements
  within a particle filter. Trained only on random interaction data,
  the model transfers
  to search and insertion trajectories on the real hardware:
  the estimator using the trained model achieves a sub-millimeter
  position error, outperforming
  the system-identified MuJoCo model. We also found that the
  predicted covariance is adaptive to the
  contact state, and improves the robustness of the estimation,
  especially when using the wrench
  measurements alone. Project website:
  \url{https://sites.google.com/view/act-fignet}.
\end{abstract}

\section{INTRODUCTION}
\label{sec:introduction}
Performing peg insertion with a wrist-mounted low resolution camera
on a robot arm faces two major challenges: the visual estimate of the
socket pose is inaccurate and more importantly, the camera is
occluded once the arm
approaches the target, making close-loop control purely with visual
clues implausible. Previous work~\cite{yi2025learning} proposed a
physics-based predictive model trained self-supervised from
proprioceptive and wrench readings, and indicated its potential role
in blind searching for the socket pose from a coarse initial guess of
the visual detection. Here we extend the method to additionally
predict contact dependent uncertainty and apply it as measurement models to
calculate likelihood for precise socket pose estimation using only
motion and wrench signals. The model is trained on the robot's
self-collected random exploration data. Nevertheless, even under
out-of-distribution probing policy, the
filter combined with the learned model still effectively reduces
uncertainty through contact and yields a sub-millimeter/degree precision for
socket pose estimation. To summarize, we make the following contributions:
\begin{itemize}
  \item an extension of the predictive model (Act-FIGNet)
    in~\cite{yi2025learning} to
    predict contact-conditional covariance of the end-effector motion
    and wrench readings, in addition to their means;
  \item a particle filter that uses the trained model as the measurement model
    for both motion and wrench to estimate the 6-D socket pose;
  \item an investigation of whether contact-adaptive and epistemic
    uncertainty benefit pose estimation.
\end{itemize}
\begin{figure}[t]
  \centering
  \includegraphics[width=\linewidth]{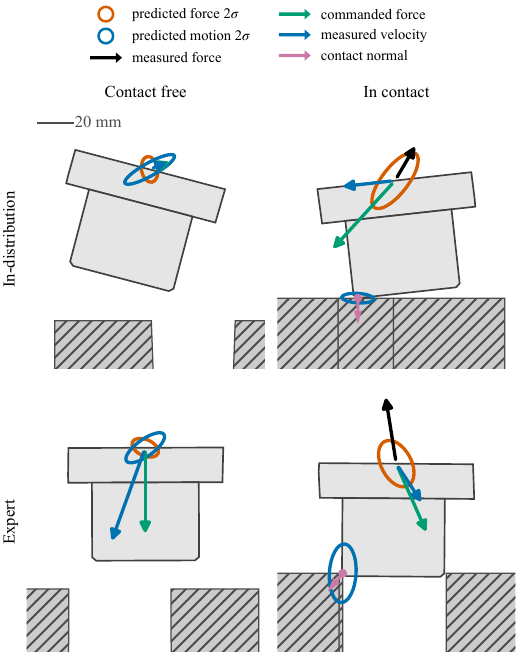}
  \caption{Shapes of the predicted covariance for motion and wrench,
    midair and in contact, under the in-distribution (ID) random
    exploration policy (\textbf{top}) and the peg insertion
    expert policy, which is out of distribution (OOD) (\textbf{bottom}). In
    contact, the motion covariance is narrow along the contact normal and
    wide in the tangent plane, and the force covariance the opposite.
    Only the position and force blocks are shown, projected onto the
  slicing plane.}
  \label{fig:cov_shape}
\end{figure}
\section{METHOD}
\subsection{Problem statement}
Given an initial belief over the socket pose,
we aim to obtain a more certain and accurate pose estimate through a series
of observed end-effector motion and wrist-mounted wrench sensor
readings. We formalize this problem as sequential Bayesian inference
of the uncertain socket pose in a partially observable Markov
decision process. We use the POMDP formulation only for consistency
with our previous work and
the eventual deployment in peg insertion; this article focuses on
inference rather than control.

\paragraph{Notation} Let $\xtheta[t]$, $\bx_t \in \mathbb{SE}(3)$ be
the socket and the robot end-effector pose at step $t$ respectively,
system state defined as $\bs_t \equiv [\xtheta[t], \bx_t,
\dot{\bx}_t]$\footnote{For our model, $\dot{\bx}_t$ is approximated
  with $\bx_t \ominus \bx_{t-1}$; for the MuJoCo baseline, which will
be introduced later, it is $(\bx_t \ominus \bx_{t-1})/\Delta t$.}. The
observable variables are the current end-effector pose and wrench
readings so measurement is defined as $\by_{t} \equiv
[\tilde{\bx}_{t}, \tilde{\bw}_{t}]$, where
$\tilde{\bw}_{t}=[\tilde{\bm{f}}_t, \tilde{\bm{\tau}}_t] \in
\mathbb{R}^6$ is the measured wrench. The measurements are predicted
by a forward model $\hat{\by}_{t}=f(\bs_{t-1}, \ba_{t-1})$. The datum
at step $t$ is $D_t=[\by_{t}, \ba_{t-1}]$. It is worth nothing that
the predicted observations here are one step ahead of the previous
state, so the filter update step will be slightly different to that
of the other filtering applications where observations are
conditioned on the current state.

Accroding to Appendix~\ref{app:bayes_inference}, the recursive update
rule for the posterior over the socket pose $\xtheta[t]$ is given by:
\begin{equation}
  \label{eq:bayesian_update}
  \begin{split}
    p(\xtheta[t] \mid D_{1:t}) \propto \bigl[\,
      & p_{\phi}(\tilde{\bx}_{t} \mid \bs_{t-1}, \ba_{t-1})\\
      \cdot\; & p_{\phi}(\tilde{\bw}_{t} \mid \bs_{t-1}, \ba_{t-1})
    \,\bigr]^{\alpha}\\
    \cdot\; & p(\xtheta[t-1] \mid D_{1:t-1})
  \end{split}
\end{equation}
where tempering factor $\alpha \in \left(0,1\right]$ accounts for the
temporal correlation between consecutive steps. The
key to a reliable estimator for the socket pose is therefore a
faithful approximation of both measurement models for motion and
wrench. In contact-rich settings, these models are non-linear and
difficult to be linearly approximated. In the following, we describe
how to approximate such non-linear models with our
contact-conditional framework (Act-FIGNet).

\subsection{Observation models}
We model the likelihood function as a Gaussian distribution:
\begin{equation}
  p_{\phi}(\by_t | \bs_{t-1}, \ba_{t-1}) = \mathcal{N}(\by_t;
  \hat{\by}_t, \hat{\bm{\Sigma}}_{\by_t})
\end{equation}
where $\hat{\by}_t=[\hat{\bx}_t, \hat{\bw}_t]$ is the predicted means
(of motion and wrench), and their covariance matrices
$\hat{\bm{\Sigma}}_{\by_t}=[\hat{\bm{\Sigma}}_{\bx_t},
\hat{\bm{\Sigma}}_{\bw_t}]$. Assuming conditional independence, the
likelihood factorizes into a motion term and a wrench term.
In the following where it causes no confusion, we omit the hat
$\hat{(\cdot)}$ on predicted quantities for brevity. Motion and
wrench distributions are predicted by the motion head
$\phi^x(\bs_{t-1},\ba_{t-1})$ and wrench head
$\phi^w(\bs_{t-1},\ba_{t-1})$ of the
learned model. The motion head actually predicts the next
end-effector acceleration $\ddot{\bx}_t \in
\mathbb{R}^6$ and its covariance $\bm{\Sigma}_{\ddot{\bx}_t} \in
\mathbb{R}^{6\times 6}$. But since the next pose is calculated with
semi-implicit Euler integration using the predicted acceleration in the form of
$\bx_t=\bx_{t-1} \oplus (\dot{\bx}_{t-1} \oplus \ddot{\bx}_t)$, the next
pose's covariance equals to that of the predicted 6-D acceleration
$\bm{\Sigma}_{\bx_t}=\bm{\Sigma}_{\ddot{\bx}_t}$. Here we denote the
addition and subtraction operations on the
$\mathbb{SE}(3)$ as $\oplus$ and $\ominus$ respectively. The Mahalanobis
distance used for motion likelihood is calculated as:
\begin{equation}
  d^2_M(\tilde{\bx}_t; \bx_t, \bm{\Sigma}_{\bx_t}) = (\tilde{\bx}_t
  \ominus {\bx}_t)^\tp\; \bm{\Sigma}_{\bx_t}^{-1}\; (\tilde{\bx}_t
  \ominus {\bx}_t)
\end{equation}
Similarly
for the Mahalanobis distance of the wrench term we have:
\begin{equation}
  d^2_M(\tilde{\bw}_t; \bw_t, \bm{\Sigma}_{\bw_t}) = (\tilde{\bw}_t -
  \bw_t)^\tp\; \bm{\Sigma}_{\bw_t}^{-1}\; (\tilde{\bw}_t - \bw_t)
\end{equation}
These two terms and their respective determinants
($|\bm{\Sigma}_{\bx_t}|$ and $|\bm{\Sigma}_{\bw_t}|$)
determine the quality of the likelihood
approximation. An ideal predictive model yields a high likelihood for hypotheses
near the ground truth and a low one elsewhere.
\subsection{Contact-conditional aleatoric uncertainty}
Conventionally, the noise covariances are tuned beforehand and fixed
during deployment. In contact-rich manipulation, however, the
prediction error varies with the contact state and its channels can
be strongly correlated; a fixed covariance can thus be overconfident
in one dimension and
uninformative in another. This motivates predicting the covariance
adaptive to the contact configuration, and investigating whether it improves
the likelihood.
Intuitively, we wish the network to learn that when contact presents,
motion along the contact normal is certainly not possible (due to
contact constraints), while in the friction-dominating plane, i.e.,
the tangent plane, it should be cautious about its prediction. Wrench
prediction is the opposite, it shouldn't be too certain along
the predicted force direction (because of controller oscillation or
sensor noise), but relatively certain in the tangent
plane, because absent contact should certainly not be causing
reaction forces other than friction that scales with the normal
force.

The framework encodes system state, applied action
and geometries including contacts into a graph, which are processed
by a stack of message-passing layers to predict the next end-effector
motion and wrench readings, decoded from the last layer node/edge
latents. To capture contact-dependent uncertainty, the decoders are
extended to predict covariance matrices. In particular, the motion
head additionally outputs an upper-triangular Cholesky factor
$\bm{L}$ of the precision
matrix ${\bm{\Sigma}}_{\bx_t}^{-1} = \bm{L}^\tp \bm{L}$, with a positive
diagonal enforced by a softplus operation. The force head cannot
predict a wrench covariance directly, since the wrench is reduced
from a per-vertex force field prediction. It
therefore predicts per-vertex mode fields, which are then reduced in the
same way as the force field, so the resulting wrench covariance is
physically coupled as well.

Implementations are detailed in
Appendix~\ref{app:uncertainty_prediction}. The training loss for both
motion and wrench heads is negative-log-likelihood of the predicted
mean and covariance. As will be shown shortly, this results in
contact adaptive and physically coupled wrench covariance shapes.

\subsection{Distance-aware epistemic uncertainty}
As with any predictive models, the learned model's calibration
degrades in the out-of-distribution domain. In the usecases of peg
insertion, the deployment state-action distribution differs
substantially from the training domain. To mitigate overconfidence on
false prediction under distribution shift, we let the network to also
quantify the epistemic uncertainty. Ensembles of the predictive
models are widely used, but their training and computation costs
scale with the number of members. Instead we adopt an OOD-distance
aware uncertainty quantification technique known as
Spectral-normalized Neural Gaussian Process
(SNGP)~\cite{DBLP:conf/nips/LiuLPTBL20}. During training, the
network's hidden layers are regulated by spectral normalization to
approximately preserve input distances in the latent space. After
training, the posterior covariance of a Gaussian-process output layer
is fitted, such that the predictive variance grows with the latent
distance to the training data. This epistemic variance is added to
the diagonal of the predicted
aleatoric covariance, so the likelihood becomes less confident as the
input moves away from the training domain. Since it is added to the
diagonal only, it weakens the predicted correlations and changes the
contact-dependent shape of the covariance. In
Section~\ref{sec:experiments}, we investigate how it changes this shape
and whether it benefits the estimation.
\begin{table}[t]
  \centering
  \caption{Contact adaptivity of the predicted (pred.) std ratio
    $\sigma_n/\sigma_t$, in the contact frame (n: contact normal, t:
    tangent plane), and of the residuals (emp.). Motion: movement at
  the contact point; force: contact force ($|f| > 1$\,N). ale.: aleatoric only.}
  \label{tab:uq_cov_physics}
  \small
  \setlength{\tabcolsep}{3pt}
  \begin{tabular}{lrrrr}
    \toprule
    & \multicolumn{2}{c}{In-distribution} & \multicolumn{2}{c}{Expert} \\
    \cmidrule(lr){2-3}\cmidrule(lr){4-5}
    & pred. & emp. & pred. & emp. \\
    \midrule
    Motion $\sigma_n/\sigma_t$ & 0.63 & 0.48 & 0.61 & 0.41 \\
    Force $\sigma_n/\sigma_t$ & 1.34 & 1.51 & 1.48 & 1.59 \\
    Motion $\sigma_n/\sigma_t$, ale. & 0.63 & 0.48 & 0.61 & 0.41 \\
    Force $\sigma_n/\sigma_t$, ale. & 1.51 & 1.51 & 1.79 & 1.59 \\
    \bottomrule
  \end{tabular}
\end{table}

\begin{figure}[t]
  \centering
  \includegraphics[width=\linewidth]{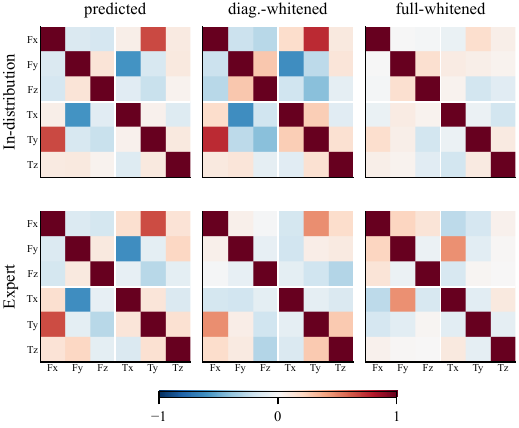}
  \caption{Correlation of the wrench in the contact frames, in
    distribution (\textbf{top}) and under the expert policy
    (\textbf{bottom}). \textbf{Left}: correlation of the predicted
    covariance. \textbf{Center}: correlation of the residual whitened
    with the diagonal of the predicted covariance only, as in the
    no-correlation variant, i.e., scaled by the predicted marginal std
    $\sigma_i$ with $\sigma_i^2=(\hat{\bm{\Sigma}}_{\bw})_{ii}$.
    \textbf{Right}: correlation of the residual whitened with the full
    predicted covariance, $\bm{L}^{-1}(\tilde{\bw}-\hat{\bw})$ with
    $\bm{L}$ the Cholesky factor of $\hat{\bm{\Sigma}}_{\bw}$. Closer
    to identity means the predicted
  correlations explain the residual coupling.}
  \label{fig:corr_heatmap}
\end{figure}

\subsection{Particle filter}
For online estimation, we adopt a particle filter, which approximates
the posterior
with a set of $N$ weighted samples $\{\bs_t^{(l)},
\omega_t^{(l)}\}_{l=1}^{N}$. At each step,
the weights are updated in log space with the tempered likelihood of
the observation:
\begin{gather}
  \ln \tilde{\omega}^{(l)}_t = \ln \omega^{(l)}_{t-1}
  + \alpha \ln p(\by_t \mid \bs_{t-1}^{(l)}, \ba_{t-1}) \\
  \omega^{(l)}_t = \frac{\tilde{\omega}^{(l)}_t}{\sum_{j}
  \tilde{\omega}^{(j)}_t}
\end{gather}
The motion and wrench likelihood terms are computed as described in the previous
section. Resampling is triggered when the effective sample size
$N_{\text{eff}} = 1 / \sum_l (\omega^{(l)}_t)^2$ drops below $0.5N$.
\paragraph{Resampling} Unlike many particle filter applications, the
process noise here is zero because the hypotheses are over the static
socket pose. In practice, we could still inject process noise to the
transition $p(\xtheta[t] | \xtheta[t-1])$ to counter particle
depletion, but the perturbed
hypotheses are very likely to violate the contact constraints that
the posterior might have implicitly
captured. Therefore,
we instead diversify the particles with a resample move
step~\cite{gilks2001following}, where new hypotheses are drawn by
a Metropolis-Hastings sampler~\cite{Hastings1970}, whose target
distribution is the current posterior $p(\xtheta[t] \mid D_{1:t})$.
Its proposal is a mixture of a Liu-West
kernel~\cite{DBLP:books/sp/01/LiuW01} and a likelihood-informed prior
proposal adapted from~\cite{cui2014likelihood,cui2016dimension}. See
Appendix~\ref{app:particle_filter} for more details.
\section{Experiment}
\label{sec:experiments}
This section aims to answer three questions by evaluating the model in
the following: (1)~whether the predicted covariance shape is
contact-adaptive (Section~\ref{sec:exp_uq}), (2)~how
well the model performs in pose estimation
(Section~\ref{sec:exp_pose}), and (3)~whether the
predicted covariance benefits the estimation
(Section~\ref{sec:exp_cov}).
\subsection{Experimental setup}
\label{sec:exp_setup}
\paragraph{Hardware} The real-world testbed is identical to that of
previous work. It comprises a UR10 arm with a wrist-mounted
force-torque sensor and a hexagonal peg, and a matching hexagonal
socket with a nominal clearance of 1~mm (per side). The arm is driven by a
Cartesian force controller at 500\,Hz whose control target is
workspace wrench commanded by the policy at 10\,Hz.
\paragraph{Training and data collection} Training follows previous
work. The data consist of joint positions and force-torque readings
collected while the robot randomly touches the socket: 450k
transitions in total, 400k for training and 50k for testing. For
evaluation only, we also collect trajectories with an \emph{expert}
policy, which first performs spiral search motion at the socket rim
then peg insertion action. This policy is OOD for the learned model,
so the evaluation requires generalization to novel state and action.
\paragraph{Baselines and ablations} We compare the learned model
against a system-identified MuJoCo model, whose simulated robot joint
and controller dynamics are identified on contact-free trajectories,
and contact solver parameters and motion/wrench noise on contact-rich
trajectories from a separate dataset. The variants for evaluation
differ in three aspects: (1)~the forward
model, Act-FIGNet or MuJoCo; (2)~the covariance, which is either the
predicted aleatoric covariance, the same with the SNGP epistemic
variance added, its diagonal only (no correlations), or a calibrated
constant covariance; and (3)~the
likelihood terms, using both, only the motion, or only the wrench
term. We additionally evaluate the predicted covariance without
calibration.
\paragraph{Calibration} All models are calibrated on 3 separate
trajectories collected with the expert policy, by maximizing the
Gaussian likelihood of the innovations at the true socket pose. For the
predicted covariance, only the scales of its linear and angular blocks
and a per-axis noise floor are fitted, so that its contact-dependent
shape is preserved; models without a predicted covariance fit a
constant diagonal covariance. We
also calculate the tempering parameter $\alpha$ via
integrated autocorrelation of the innovations~\cite{ribatet2012bayesian,
geyer1992practical}. Calibration details are given in
Appendix~\ref{app:calibration}.
\subsection{Contact adaptivity of the predicted covariance}
\label{sec:exp_uq}
To evaluate the contact adaptivity of the predicted covariance, we
examine its shape and correlation, but not its scale, which the
calibration fits while the shape remains adaptive. The evaluation uses
the data frames in which the tool is in contact. The motion covariance is
predicted for the tool pose in the tool's body frame, so it is first
transformed to the contact point in the contact frame,
$\bm{J}_c \bm{\Sigma} \bm{J}_c^\tp$ with $\bm{J}_c =
\bm{R}_c\bigl[\bm{I}_3,\; -[\mathbf{p}_c]_{\times}\bigr] \in
\mathbb{R}^{3 \times 6}$, where $\bm{R}_c$ is the rotation of the
contact frame and $\mathbf{p}_c$ points from the body frame's origin to
the contact point; the force covariance is only rotated by $\bm{R}_c$.
In contact, the motion should be more certain along the contact normal
than in the tangent plane, since the contact constrains the normal
motion. We quantify this with the ratio $\sigma_n/\sigma_t$ of the
standard deviations along the normal and in the tangent plane, which
should be smaller than 1. For the force, the opposite is expected: the
oscillation of the force controller acts on the normal force, while
friction bounds the tangential force to a fraction of it, so the ratio
should be larger than 1.

Table~\ref{tab:uq_cov_physics} shows that the predicted ratios lie on
the expected side of 1 for both modalities, in distribution as well as
under the OOD expert policy. The predicted
motion covariance underestimates the anisotropy of the residuals,
while the aleatoric force covariance matches it in distribution. Since
the epistemic variance is added to the diagonal, it makes the force
covariance more isotropic, whereas the motion covariance remains
unchanged by it. Figure~\ref{fig:cov_shape} illustrates these shapes: wide
in the tangent plane and narrow along the contact normal for motion,
and the opposite for force. The complementary covariance shapes of the two
modalities is also visible; each is confident in the directions
where the other is uncertain. Fig.~\ref{fig:corr_heatmap} shows the
cross-dimensional correlation of the predicted wrench covariance. Its
dominant entries couple the lateral forces with the torques about the
orthogonal axes, which reflects the lever arm of the contact force. In
distribution, the residuals show the same coupling, and whitening them
with the predicted full covariance removes most of the correlation.
Under the expert policy, the residual coupling is weaker than
predicted, so the full whitening over-corrects it, possibly because
the peg insertion trajectories contain mostly small torque readings.

\subsection{Socket pose estimation}
\label{sec:exp_pose}
\paragraph{Protocol} Previous experiments demonstrate
the prediction quality via contact-adaptivity evaluation. Here we
test the models with end-to-end pose estimation runs using a particle
filter. The filter is run on 10 real insertion trajectories recorded
with the expert policy. The same trajectory is run with three random
seeds; for our learned model, we further evaluate models trained with
three different seeds, making 90 runs for the learned baselines and
30 for the MuJoCo baseline. The prior is a Gaussian in the tangent
space at the true socket pose
$\xtheta[0] = \bm{x}^{\theta}_{\text{gt}} \oplus \bm{\epsilon}$,
whose mean is itself randomly biased:
\begin{equation}
  \bm{\epsilon} \sim \mathcal{N}(\bm{\mu}_{\text{prior}},
  \bm{\Sigma}_{\text{prior}}) \qquad
  \bm{\mu}_{\text{prior}} \sim \mathcal{N}(\bm{0}, \bm{\Sigma}_{\text{bias}})
\end{equation}
We use $\bm{\Sigma}_{\text{prior}} = \operatorname{diag}(0.01^2\,
\mathbf{1}_{3},\; 0.05^2\,\mathbf{1}_{3})$, i.e., standard deviations
of 10\,mm for the translations and 0.05\,rad (\ang{2.9}) for the
rotations, and a lateral bias with a standard deviation of 10\,mm in
$x$ and $y$ ($\bm{\Sigma}_{\text{bias}}$ is zero otherwise). The filter
uses $N = 200$ particles.

\begin{figure*}[t]
  \centering
  \begin{subfigure}{0.49\textwidth}
    \centering
    \includegraphics[width=\linewidth]{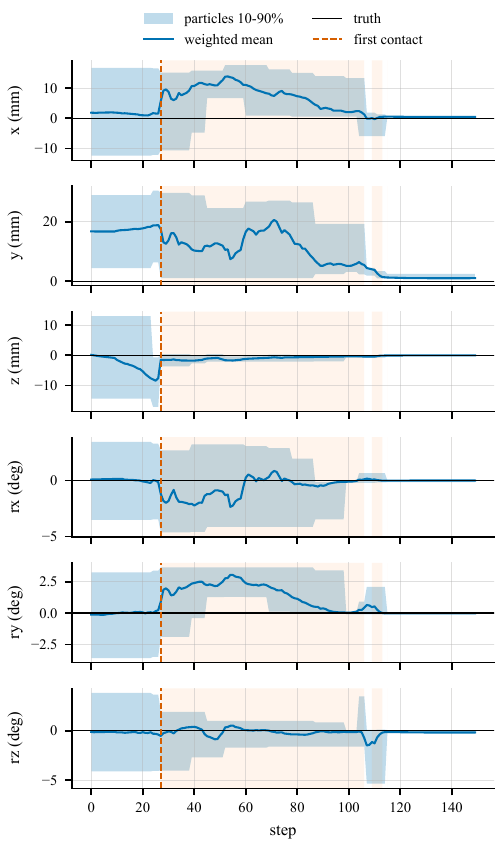}
    \caption{Act-FIGNet.}
  \end{subfigure}
  \hfill
  \begin{subfigure}{0.49\textwidth}
    \centering
    \includegraphics[width=\linewidth]{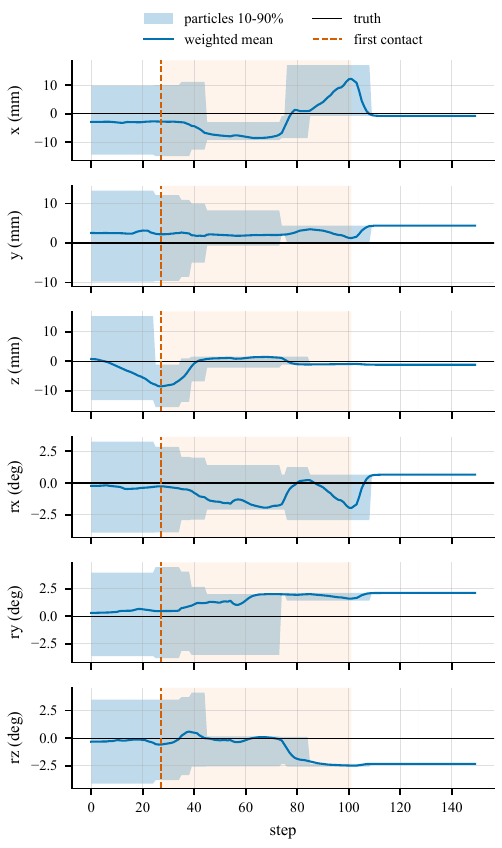}
    \caption{System-identified MuJoCo.}
  \end{subfigure}
  \caption{Particle convergence in 6-D socket pose estimation. With both
    models, the estimate along $z$ converges at the first
    contact, and with the learned model, the lateral estimate converges
  once the peg enters the socket.}
  \label{fig:estimate_full_noise}
\end{figure*}

The expert policy collects probing
trajectories, because they include the insertion, and hence provide
sufficient evidence to constrain the full 6-D socket pose. The
filter is not informed of the
insertion and has to
infer the socket pose from motion and wrench alone.
Table~\ref{tab:run_report} reports the filtering
accuracy as the final lateral, tilt errors, and errors along and
around the z-axis, because they are the
quantities used in the MPC
objective of our previous work~\cite{yi2025learning} and thus directly
reflect the insertion success. For each, we
report the median and the percentage of runs within a threshold:
1\,mm for the lateral position, which equals the clearance, and $z$,
and \ang{2} for tilt and rotation around $z$, given the socket
clearance and the peg length of
45\,mm.
\paragraph{Results} With both measurement terms, our model without
calibration already converges
within the lateral
and tilt thresholds in 98\% and 99\% of the runs, against 37\% and 53\% for
MuJoCo. The estimate with the scaled predicted
covariance converges
at similar rates. With a constant covariance, the estimator with the
contact-conditional mean also performs well, reaching 93\% and 100\%.
With both terms, all variants are more robust than MuJoCo, consistent
with~\cite{yi2025learning} that our model predicts
motion and wrench more accurately.
Fig.~\ref{fig:estimate_full_noise} shows the convergence of particles
over an example trajectory. For this particular search then insert
pattern, estimate over the z-axis converges to the ground truth the
moment it touches the surface, while the spiral search on the
rim constrains the lateral position only weakly. Once the peg enters
the socket, the lateral estimate also converges, since any hypotheses
with lateral errors larger than the
clearance will impede the motion at the rim, which contradicts the
observed insertion, and hence receive low likelihood.
This demonstrates the learned measurement models correctly refute
physically implausible hypotheses. Since the model is trained only on
random contact-rich interactions, this
also shows that its predictions transfer to the unseen search and
insertion motions.

\subsection{Ablation study}
\label{sec:exp_cov}
\begin{figure}[t]
  \centering
  \includegraphics[width=\linewidth]{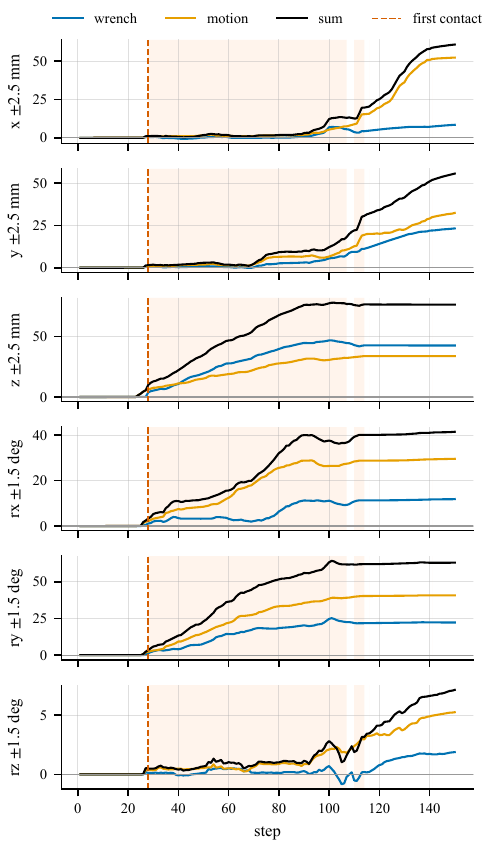}
  \caption{Accumulated log-likelihoods of the wrench and motion terms
    at the ground truth $\sum_t\ln p_\phi(\tilde{\bw}_t,
    \tilde{\bx}_t|\xtheta)$ against at the perturbed hypotheses
    $\sum_t\ln p_\phi(\tilde{\bw}_t, \tilde{\bx}_t|\xtheta \oplus
    \bm{\epsilon})$ over a 6-D pose estimation trajectory. Perturbation
    for position $\epsilon_{\text{pos}}=\pm 2.5\text{mm}$ and for
  orientation $\epsilon_{\text{ori}}=\pm \ang[round-mode=none]{1.5}$.}
  \label{fig:loglik_profile}
\end{figure}

\paragraph{Modalities} Using only the wrench term, none of the models
converges reliably in the lateral position. Interestingly, the fitted
scales negatively affect the wrench term, whose convergence rate
drops from 67\% to 41\% for lateral and from 89\% to 71\% for $z$
estimation. Using only the motion
term, the variants converge more reliably, with the rates ranging
from 84\% to 94\%. Combining both terms leads to higher rates than
each term alone, suggesting they are complementary in our model. For
MuJoCo, adding the wrench term instead degrades the estimate along
$z$ and in tilt, but slightly improves convergence in the lateral
plane. A possible explanation is that the identified contact
parameters trade wrench for motion accuracy, leading to inconsistent
predictions of these two. To
further examine the contribution of
each term, we take a random episode, then perturb the socket pose
away from the ground truth, and compare the accumulative wrench and motion
likelihoods along the trajectory relative to those at the true pose.
Fig.~\ref{fig:loglik_profile} shows the result: for lateral position
and orientation, the motion term has more influence on the total
likelihoods, especially along the $x$-axis, where the motion term
dominates in this episode. Along the $z$-axis, the wrench term shapes
the likelihoods more strongly.

\paragraph{Covariance} With both (motion and wrench) terms, the predicted
aleatoric covariance without scaling already achieves the same convergence rate
as the calibrated ones. Removing the predicted correlations visibly lowers the
rates when each term is used alone, especially for the wrench term;
these rates rank the lowest among the variants with predicted and
calibrated covariance. With both terms combined, however, the rates
are almost unaffected. So the correlations are crucial for each term
alone, but combining both compensates for their absence.
The variant with a calibrated noise covariance
attains comparable rates as the ones with predicted covariance. This
suggests that, when using both terms, the contact-conditional mean is
already sufficiently discriminative even under distribution shift, or
that the evidence in these
trajectories is too abundant to reveal a difference. This might also
explain why inflating the covariance with the epistemic variance
brings no visible benefit. However, the contact-adaptive covariance
substantially improves the
robustness when the less informative wrench term is used alone; without
calibration, it raises the convergence rate over the calibrated
constant covariance from 23\% to 67\% in the lateral position and from
26\% to 89\% along $z$.

Appendix~\ref{app:particle_filter} further examines the likelihood
distribution of these
variants as a function of $\xtheta$ near the ground truth (see
Fig.~\ref{fig:likelihood_map}), and it shows that the likelihood of every
variant peaks close to the true pose, also with the wrench term
alone. It is likely that the wrench-only failures mostly come from
the sampling. Because during the early contact phase, before
insertion, the wrench term is barely informative (see
Fig.~\ref{fig:loglik_profile}), so the particles can settle in a
wrong mode and fail to recover from it afterwards.

\subsection{Runtime}
One filter step with 200 particles takes roughly
100\,ms on the GPU, but the resampling step takes longer, since
it replays the trajectory up to the current step and its runtime scales with the
history length. A fix sliding window approximation or running the
resampling step in the background with a separate GPU would reduce
the runtime, implementation of which is left for future work.

\begin{table}[t]
  \centering
  \caption{Final-step socket pose error, median and percentage of
  episodes within the threshold over episodes, model seeds and filter seeds.}
  \label{tab:run_report}
  \footnotesize
  \setlength{\tabcolsep}{3.5pt}
  \begin{tabular}{lrcrcrcrc}
    \toprule
    & \multicolumn{2}{c}{Lat. (mm)} & \multicolumn{2}{c}{z (mm)} &
    \multicolumn{2}{c}{Tilt ($^\circ$)} & \multicolumn{2}{c}{Rot. z
    ($^\circ$)} \\
    \cmidrule(lr){2-3}\cmidrule(lr){4-5}\cmidrule(lr){6-7}\cmidrule(lr){8-9}
    & med. & \% $\le$1 & med. & \% $\le$1 & med. & \% $\le$2 & med. &
    \% $\le$2 \\
    \midrule
    \multicolumn{9}{l}{\emph{Act-FIGNet}} \\
    Both & 0.54 & 98 & 0.23 & 100 & 0.16 & 99 & 0.31 & 100 \\
    Wrench & 1.10 & 41 & 0.44 & 71 & 0.51 & 90 & 0.39 & 98 \\
    Motion & 0.53 & 90 & 0.23 & 100 & 0.22 & 94 & 0.40 & 100 \\
    \midrule
    \multicolumn{9}{l}{\emph{Act-FIGNet, unscaled}} \\
    Both & 0.60 & 98 & 0.23 & 99 & 0.17 & 99 & 0.29 & 100 \\
    Wrench & 0.86 & 67 & 0.31 & 89 & 0.41 & 87 & 0.46 & 97 \\
    Motion & 0.57 & 84 & 0.30 & 97 & 0.35 & 99 & 0.51 & 99 \\
    \midrule
    \multicolumn{9}{l}{\emph{Act-FIGNet, no correlations}} \\
    Both & 0.62 & 96 & 0.24 & 99 & 0.18 & 100 & 0.33 & 100 \\
    Wrench & 1.64 & 27 & 0.85 & 53 & 0.91 & 88 & 0.43 & 96 \\
    Motion & 0.65 & 84 & 0.24 & 99 & 0.43 & 96 & 0.49 & 100 \\
    \midrule
    \multicolumn{9}{l}{\emph{Act-FIGNet + SNGP}} \\
    Both & 0.55 & 97 & 0.23 & 99 & 0.16 & 99 & 0.32 & 100 \\
    Wrench & 1.15 & 43 & 0.39 & 66 & 0.56 & 90 & 0.34 & 98 \\
    Motion & 0.54 & 94 & 0.21 & 100 & 0.23 & 97 & 0.33 & 100 \\
    \midrule
    \multicolumn{9}{l}{\emph{Act-FIGNet + constant covariance}} \\
    Both & 0.63 & 93 & 0.21 & 100 & 0.19 & 100 & 0.45 & 100 \\
    Wrench & 2.33 & 23 & 6.77 & 26 & 0.67 & 81 & 0.30 & 99 \\
    Motion & 0.64 & 88 & 0.16 & 100 & 0.22 & 99 & 0.59 & 99 \\
    \midrule
    \multicolumn{9}{l}{\emph{MuJoCo + constant covariance}} \\
    Both & 2.38 & 37 & 1.34 & 37 & 1.49 & 53 & 0.18 & 83 \\
    Wrench & 2.07 & 27 & 1.91 & 27 & 2.38 & 47 & 0.23 & 80 \\
    Motion & 1.88 & 20 & 0.31 & 97 & 0.75 & 100 & 0.27 & 100 \\
    \bottomrule
  \end{tabular}
\end{table}

\begin{figure}[t]
  \centering
  \includegraphics[width=\linewidth]{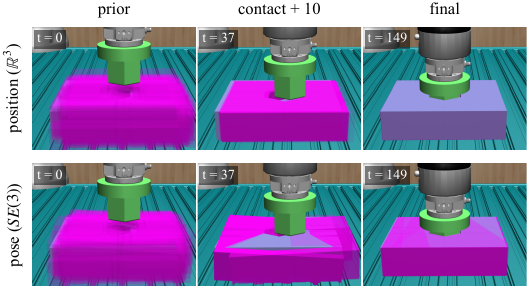}
  \caption{Snapshots of socket pose estimation with Act-FIGNet along an
    example trajectory. The socket pose hypotheses (particles) are shown
    as purple overlays.
    \textbf{Top}: Estimation with known orientation. \textbf{Bottom}:
  Full 6-D pose estimation.}
  \label{fig:snapshots_gnn}
\end{figure}

\section{Related work}

\subsection{State estimation} \cite{DBLP:conf/icra/WirnshoferSMWB19}
use analytical expressions to solve the contact dynamics in closed
form, and compute likelihoods from the predicted and measured joint
positions. They further argue that a motion-based measurement model
has an advantage over a wrench-based one, because accurate wrench
predictions on real robots are hard to obtain without careful system
identification. However, they assume a joint-torque-controlled robot,
which is not always available, especially among low-cost robots.
\cite{DBLP:conf/iros/Pankert023} compute likelihoods from forward
kinematics and binary contact measurements, and combine them with an
uncertainty-reducing policy for in-hand object pose estimation.
\cite{DBLP:journals/corr/abs-2602-19651} directly learn the dynamics
and observation models for particle filtering, but these models are
not contact-conditioned.

\section{Discussion}
This work extends previous method to predict contact-conditional
uncertainty and applies it to socket pose estimation. A particle
filter using the learned model as measurement model achieves high
accuracy with only proprioceptive and haptic clues. This provides an
effective way to refining the rough estimate given by the visual
perception.

However, the evaluation here is limited to pre-recorded
trajectories that include the
insertion. The reported accuracy thus only reflects the measurement model
given sufficient evidence; the accuracy attainable would
depend on the probing policy and is not evaluated here.

We also
found that the predicted covariance adapts to the
contact state and improves the estimation convergence, most when only
the wrench measurements are available. On the recorded peg insertion
trajectories, the filter with a constant covariance already performs
well; the benefits of the contact-adaptive uncertainty might become
more evident under more challenging probing trajectories. A natural
next step is to
couple the filter with an uncertainty-informed
policy, using the same model as the dynamics model for MPC and as the
measurement model for inference.
Instead of optimizing for the task
objective, the MPC from the prior work can be guided by the belief of
the current state to maximize the information gain, which would extend
our framework to an interactive perception pipeline.

As in our
previous work, the method has been evaluated on a single
robot and a single peg-socket pairs only. Evaluation across robots,
controllers, object materials and geometries is left for future work.

\section*{APPENDIX}

\subsection{Recursive Bayesian inference}
\label{app:bayes_inference}

Since the trajectories are pre-recorded, we treat the actions as given
inputs and ignore their dependence on the socket pose.
Given a sequence of data $D_{1:t}$ the posterior of $\bs_t$ is:

\begin{align*}
  p(\bs_t \mid D_{1:t})
  = \int p(\bs_t \mid \bs_{t-1}, D_{1:t})\;
  p(\bs_{t-1} \mid D_{1:t})\, d\bs_{t-1}
\end{align*}

The first term is the transition distribution. With observed
end-effector pose and assuming that the socket is fixed, belief over
$\bs_t$ factorizes into a distribution over the socket pose a point
mass on the observed pose:
\begin{align}
  \begin{split}
    p(\bs_t \mid \bs_{t-1}, D_{1:t})
    &= p(\bs_{t} \mid \bs_{t-1}, \by_t, \ba_{t-1})\\[0.5ex]
    &= \underbrace{p(\xtheta[t] \mid \xtheta[t-1])}_{\text{identity}}
    \cdot \underbrace{p(\bx_{t} \mid
    \tilde{\bx}_t)}_{\delta(\bx_t-\tilde{\bx}_t)}
  \end{split}
\end{align}
However, the static socket assumption is not strictly imposed; if the
socket is moving as an effect of the end-effector's motion, this term
just needs to adjust accordingly (potentially also predicted by the
same forward model).

The second term is the evidence update:
\begin{equation}
  \begin{split}
    p(\bs_{t-1} \mid D_{1:t})
    &= p(\bs_{t-1} \mid \by_{t}, \ba_{t-1}, D_{1:t-1})\\
    &= \eta \; p(\by_{t} \mid \bs_{t-1}, \ba_{t-1})\\
    &\quad\ \cdot\, p(\bs_{t-1} \mid D_{1:t-1})
  \end{split}
\end{equation}
Likelihood is approximated assuming wrench and motion observations
are conditionally independent:
\begin{equation}
  \begin{split}
    p(\by_{t} \mid \bs_{t-1}, \ba_{t-1})\, \approx \;
    & p_{\phi}(\tilde{\bx}_{t} \mid \bs_{t-1}, \ba_{t-1})\\
    \cdot\; & p_{\phi}(\tilde{\bw}_{t} \mid \bs_{t-1}, \ba_{t-1})
  \end{split}
\end{equation}
Although it seems strange that the observed end-effector pose
$\tilde{\bx}_t$ appears both in the transition and the evidence
update, it plays different roles in those terms. In the transition
update it enters as $\delta{(\bx_t-\tilde{\bx}_t)}$ to ensure that
the next likelihood is calculated from the observed $\bx_t$, while
serving as a piece of evidence for the hypothesis on the previous
state in the likelihood term. This is due to the fact that both the
motion and wrench readings are predicted by a single forward model.

Finally, the socket pose posterior is retrieved by marginalizing the
full state posterior:
\begin{equation}
  p(\xtheta[t] \mid D_{1:t}) = \int p(\bs_t \mid D_{1:t}) \,d\bx_t
\end{equation}
Because the $\bx_t$ component of $\bs_t$ is observed, whose
distribution is just a point mass given the observed trajectory
$\tilde{\bx}_{1:t}$, it results in the update rule in
Equation~\eqref{eq:bayesian_update}.

\subsection{Particle filter}
\label{app:particle_filter}
\begin{figure*}[t]
  \centering
  \begin{subfigure}{0.49\textwidth}
    \centering
    \includegraphics[width=\linewidth]{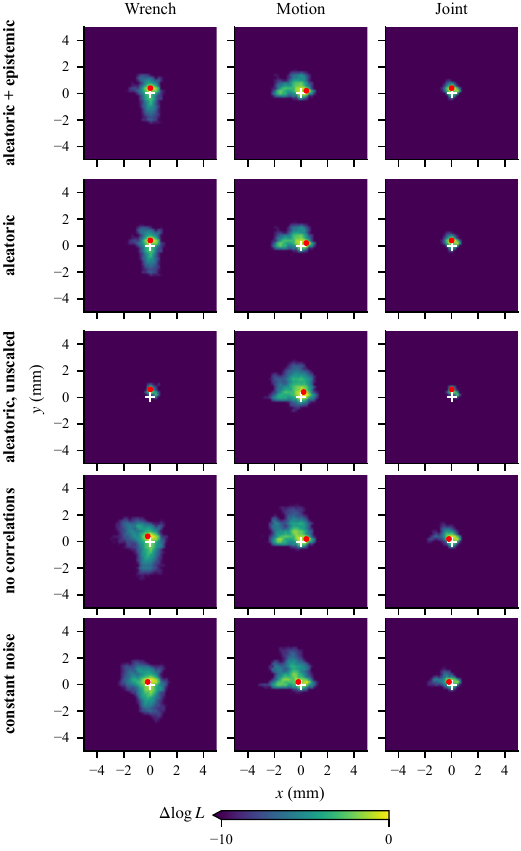}
    \caption{Lateral position.}
    \label{fig:likelihood_map_xy}
  \end{subfigure}
  \hfill
  \begin{subfigure}{0.49\textwidth}
    \centering
    \includegraphics[width=\linewidth]{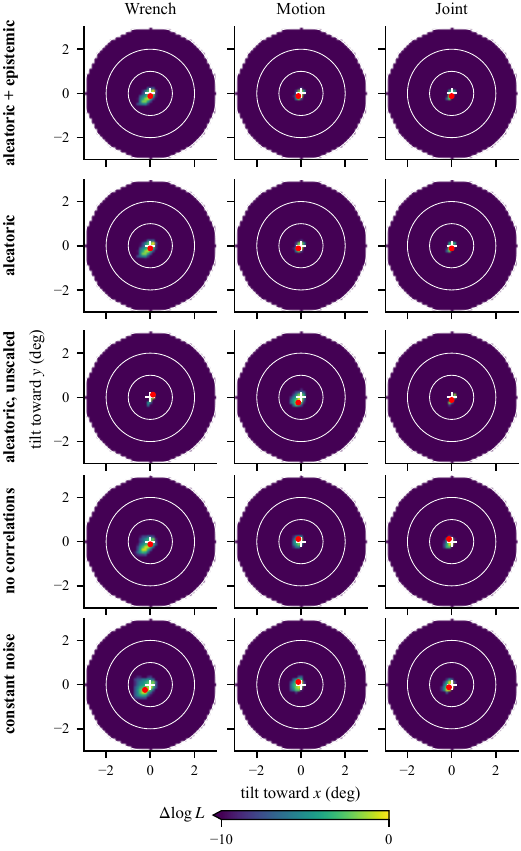}
    \caption{Tilt.}
    \label{fig:likelihood_map_lean}
  \end{subfigure}
  \caption{Likelihood maps of the covariance variants in
    Table~\ref{tab:loglik_eval} around the ground truth (white cross)
    for an example episode at step 150, computed as in the table.
    Likelihood-weighted means
  are drawn as red dots.}
  \label{fig:likelihood_map}
\end{figure*}

\paragraph{Metropolis-Hastings sampling}
After resampling, the equally weighted particles approximate the
posterior $p(\xtheta[t] | D_{1:t})$. Since the socket pose
$\xtheta[t]$ is static, there is no process noise to move the
particles, and repeated resampling leads to sample impoverishment. We
could inject independent artificial noise to preserve diversity, but
it would discard important information, for example the perturbed new
samples would violate contact constraints that the target
distribution might have already captured. We therefore adopt a
resample-move scheme~\cite{gilks2001following}: after resampling,
each particle undergoes a Metropolis-Hastings
step~\cite{Hastings1970} whose target distribution is the current
posterior. The proposal $q$ is a mixture of the Liu-West
kernel~\cite{DBLP:books/sp/01/LiuW01} and the likelihood-informed
prior/subspace (LIS)~\cite{cui2014likelihood}. The Liu-West
kernel perturbs particles locally, shrinking toward the particle
mean, while the LIS allows guided global moves,
helping to escape from wrong modes. Together the proposal is:
\begin{equation}
  q = \beta\, q_{\text{LW}} + (1-\beta)\, q_{\text{LIS}}
\end{equation}
where $\beta \in [0,1]$ is the mixture weight, and we here use
$\beta=0.5$. Since this proposal is asymmetric, the acceptance probability is:
\begin{equation}
  \rho = \min\!\left(1,\;
    \frac{p(\xtheta' \mid D_{1:t})\, q(\xtheta \mid \xtheta')}
    {p(\xtheta \mid D_{1:t})\, q(\xtheta' \mid \xtheta)}
  \right)
  \label{eq:proposal}
\end{equation}
The target distribution is evaluated over the full history, and for a
static socket pose, this becomes:
\begin{equation}
  \begin{split}
    p(\xtheta \mid D_{1:t}) \;\propto\;
    & \prod_{k=1}^{t} p_\phi\bigl(\by_k \mid \bs_{k-1}(\xtheta),\,
    \ba_{k-1}\bigr)^{\alpha}\\
    \cdot\;& p_{\mathrm{prior}}(\xtheta)
  \end{split}
\end{equation}
where $\bs_{k-1}(\xtheta) = [\xtheta,\, \tilde{\bx}_{k-1},\,
\dot{\tilde{\bx}}_{k-1}]$. The evaluation requires running the
forward model on all $t$ past transitions
for every proposal. To save computation, a proposed sample is
rejected as soon as its partial
log-likelihood falls below the acceptance threshold\footnote{Since the
  calibrated covariance is bounded from below by the additive noise
  floor, $\bm{\Sigma} \succeq
  \operatorname{diag}(\boldsymbol{\sigma}_{\min})^{2}$, each per-step
log-likelihood is bounded from above.}. Proposals that contradict
past motion and wrench
observations will be rejected, preserving the information (e.g.,
implicit contact constraints) accumulated so far while preventing
impoverishment. Since the history is fixed, these evaluations are
conditionally independent of each other so they can be run
efficiently on GPU by batching across particles and time steps (given
enough device memory, this corresponds to only one forward pass). We
use four MH move steps for every variant.
\paragraph{Likelihood} The accumulated log-likelihood along the
trajectory, $\alpha\sum_{k} \ln p_\phi(\by_k \mid \bs_{k-1}(\xtheta),
\ba_{k-1})$, determines how accurate the final estimate can be: if it
favors a wrong mode over the true $\xtheta$, the estimate will be
biased. We therefore evaluate it on grids through the ground truth in
the lateral and tilt dimensions, and report the bias and spread of the
resulting likelihood distribution in Table~\ref{tab:loglik_eval}, for
the predicted covariance with and without the SNGP epistemic variance,
for its diagonal only (no correlations), and for the calibrated
constant covariance. All variants peak close to the ground truth with a
lateral bias roughly between 0.5 and 0.7\,mm, comparable to the final filter
error in Table~\ref{tab:run_report}.
Fig.~\ref{fig:likelihood_map} visualizes the likelihood maps.
\paragraph{Liu-West kernel} The proposal is given by:
\begin{equation}
  q_{\text{LW}}(\xtheta' \mid \xtheta) =
  \mathcal{N}\bigl(\xtheta';\; \bar{\bx}^{\theta} \oplus
  a\,(\xtheta \ominus \bar{\bx}^{\theta}),\; h^2 \bm{V}\bigr)
  \label{eq:liu_west_kernel}
\end{equation}
where $\bar{\bx}^{\theta}$ and $\bm{V}$ are the mean and covariance
of the particle set, $h \in (0,1)$ is the kernel smoothing parameter,
and $a = \sqrt{1-h^2}$ is the shrinkage factor that preserves the
particle covariance. To ensure exploration even when the particles
collapse, $\bm{V}$ is regularized by
$\operatorname{diag}(\bm{\sigma}_{\text{LW}}^2)$, which makes the
covariance preservation approximate. We use $h = 0.5$ and standard
deviations $\bm{\sigma}_{\text{LW}}$ of 0.5\,mm for the translations
and 0.1\,rad for the rotations.
\paragraph{Likelihood-informed prior proposal} Sampling all dimensions
from the prior is inefficient with the $N=200$ particles used here, because
the likelihood is peaked around the ground truth, so most prior
samples are rejected. Following the likelihood-informed
subspace~\cite{cui2014likelihood}, we therefore draw from the prior
mainly in the weakly observed directions. In our setup, for instance,
the position along the $z$-axis is strongly observed, whereas the
rotation around it is weakly observed. To identify these directions,
we compute the central-difference Jacobian of the innovation at the
belief mean $\bar{\bx}^{\theta}$ for each likelihood term in use:
\begin{equation}
  \bm{H}_{:,i} = \bm{\nu}\left(\bar{\bx}^{\theta} \oplus \epsilon_i
  \mathbf{e}_i\right)
  - \bm{\nu}\left(\bar{\bx}^{\theta} \ominus \epsilon_i \mathbf{e}_i\right)
\end{equation}
where $\bm{\nu}(\xtheta) \in \mathbb{R}^6$ is the difference between
the predicted and observed measurements as a function of $\xtheta$,
$\mathbf{e}_i$ the $i$-th unit vector, and
$\epsilon_i=0.5\,\sigma_{\text{prior},i}$ the step length, with
$\sigma_{\text{prior},i}$ the prior standard deviation of dimension
$i$. Since $2\epsilon_i = \sigma_{\text{prior},i}$, $\bm{H}$ is the
Jacobian in units of the prior standard deviation. Treating the
covariance as constant\footnote{For the models with a predicted
  covariance, $\bm{\Sigma}$ varies with $\xtheta$; we neglect this
dependence for simplicity.}, the Fisher information of each likelihood term in
use is accumulated over the same history as the MH target:
\begin{equation}
  \bm{F} = \sum_{k=1}^{t} \bm{H}_k^\tp \bm{\Sigma}_k^{-1} \bm{H}_k
\end{equation}
where $\bm{H}_k$ is the Jacobian of the innovation at step $k$, and
with both terms, their information matrices are added.
Its eigenvectors give the principal directions, and its eigenvalues
compare the information of the untempered likelihood in each
direction with that of the prior, as in~\cite{cui2014likelihood}.
Along directions with
small eigenvalues, the proposal draws from the prior; directions with
large eigenvalues are drawn from the prior with a lower probability.
Unlike the hard split into subspaces in~\cite{cui2014likelihood}, the
eigenvalues set the probability of drawing from the prior. We
denote this proposal by $q_{\text{LIS}}$. In our setup, the position
along $z$ is resampled less often than that along $x$ and
$y$, because $z$ is strongly observed in comparison.
\subsection{Aleatoric uncertainty prediction}
\label{app:uncertainty_prediction}
The framework encodes state, action, object geometries, and contact
state into a graph, which is then passed through a stack of
message-passing steps. Motion and wrench predictions are decoded from
the final-layer latent graph. The motion head decodes the object node
latent $\bm{v}^{O}$ while the wrench head decodes the wrench-vertex
edge latent $\bm{e}_{m_i \to w}$.

The motion head produces object-level motion prediction directly:
\begin{gather}
  \bigl[\hat{\ddot{\bx}}_t, \bm{L}\bigr]=\phi^{\bx}(\bm{v}^{O})
\end{gather}
The wrench head yields per-vertex force and mode fields:
\begin{gather}
  \bigl[\, \hat{\bm{f}}_i,\; b_i,\; \bm{m}^{1}_i, \ldots,
  \bm{m}^{K}_i \,\bigr]_{i}^{\mathcal{N}} = \phi^{\bw}(\bm{e}_{m_i \to w})
\end{gather}
where $b_i$ is a scalar and $\bm{m}^k_i \in \mathbb{R}^3$ the mode
fields, with $i$ the index of $\mathcal{N}$ vertices; the first mode
field is tied to the predicted force field
$\bm{m}^{0}_i = b_i\, \hat{\bm{f}}_i$, so that it is parallel to the
predicted force at each vertex. Let $\langle\bm{q}\rangle \equiv
\frac{1}{\mathcal{N}}\sum_{i}\bm{q}_i$
be the mean of a per-vertex quantity, and $\mathring{\bm{q}}_i \equiv \bm{q}_i -
\langle\bm{q}\rangle$ the centered quantity. The force and mode
fields can then be reduced to object-level the same way using the
object's geometry:
\begin{gather}
  \hat{\bm{w}}_t =
  \begin{bmatrix}
    \langle \hat{\bm{f}} \rangle \\[2pt]
    \bm{G}^{-1}\big\langle \mathring{\bm{r}} \times \hat{\bm{f}} \big\rangle
  \end{bmatrix}
  \qquad
  \hat{\bm{u}}_k =
  \begin{bmatrix}
    \langle \bm{m}^{k} \rangle \\[2pt]
    \bm{G}^{-1}\big\langle \mathring{\bm{r}} \times \bm{m}^{k} \big\rangle
  \end{bmatrix}
\end{gather}
with $\bm{G} = \langle\bm{P}\rangle$, $\bm{P}_i = \bm{I} -
\mathring{\bm{r}}_i \mathring{\bm{r}}_i^\tp /
\lVert\mathring{\bm{r}}_i\rVert^2$, and $\mathring{\bm{r}}_i$ is the
vertex lever-arm around the centroid. The full wrench covariance
matrix is calculated from the $K+1$ mode fields:
\begin{gather}
  \hat{\bm{\Sigma}}_{\bm{w}} = \sum_{k=0}^{K} \hat{\bm{u}}_k
  \hat{\bm{u}}_k^{\!\tp}
  + \bm{\Sigma}_{\text{base}}
\end{gather}
where $\bm{\Sigma}_{\text{base}} =
\bm{L}_{\text{base}}\bm{L}_{\text{base}}^\tp \in \mathbb{R}^{6 \times
6}$, with a learnable Cholesky factor $\bm{L}_{\text{base}}$, ensures
that the covariance is positive definite and invertible. The $K+1$
mode fields give the covariance its contact-adaptive shape. We use
$K=3$, so the adaptive part has rank at most $K+1=4$ and spans at most
four directions, while $\bm{\Sigma}_{\text{base}}$ provides the full
rank. Multiple modes are needed to capture the torque variation, because with
$K=0$, the adaptive part would be of rank one, so its torque variation
would be fully determined by its force variation.
\subsection{Calibration}
\label{app:calibration}
\paragraph{Observation noise} The observation noise of each term is
fitted by maximizing the Gaussian likelihood of the innovations at the
true pose over the contact steps of the calibration trajectories. For
models with a predicted covariance $\bm{\Sigma}$, the filter uses
$\bm{S}\,\bm{\Sigma}\,\bm{S} +
\operatorname{diag}(\bm{\sigma}_{\min})^{2}$, where $\bm{S} =
\operatorname{diag}(s_{1}\mathbf{1}_{3},\; s_{2}\mathbf{1}_{3})$ with
$s_1, s_2 > 0$ scales the linear and angular blocks (force/torque for
the wrench, position/rotation for the motion), and
$\bm{\sigma}_{\min} \in \mathbb{R}^{6}_{+}$ is a per-axis noise floor
that keeps the covariance positive definite and accounts for unmodeled
sensor noise. The scales and the floor are fitted with the Nelder-Mead
method~\cite{DBLP:journals/cj/NelderM65, 2020SciPy-NMeth}, starting
from $s_1, s_2$ set to the
block RMS of the innovations normalized by the predicted standard
deviations. For models without a predicted covariance, the filter
uses $\operatorname{diag}(\bm{\sigma}_{\text{noise}})^{2}$, whose
maximum-likelihood solution is the per-axis RMS of the innovations.

\paragraph{Tempering factor} As in composite-likelihood
adjustments~\cite{ribatet2012bayesian}, the tempering factor is set to $\alpha =
1/\tau_{\text{int}}$, where
$\tau_{\text{int}}$ is the integrated autocorrelation time of the
residual at the true pose. Let $\nu_{j,t,i}$ be the $i$-th component
of the whitened innovation at step $t$ of contact segment $j$ with
length $T_j$, and $\mathring{\nu}_{j,t,i} = \nu_{j,t,i} -
\bar{\nu}_{j,i}$ its value centered within the segment. The
autocorrelation at lag $k$, averaged over the axes, is
\begin{equation}
  \gamma_k = \frac{1}{\dim\bm{\nu}}\sum_{i=1}^{\dim\bm{\nu}}
  \frac{\sum_{j}\sum_{t=1}^{T_j-k}
  \mathring{\nu}_{j,t,i}\,\mathring{\nu}_{j,t+k,i}}
  {\sum_{j}\sum_{t=1}^{T_j}\mathring{\nu}_{j,t,i}^{2}}
\end{equation}
The integrated autocorrelation time is then
\begin{equation}
  \tau_{\text{int}} = 1 + 2\sum_{k=1}^{k^{*}-1} \gamma_k
  \qquad
  k^{*} = \min\{k \geq 1 : \gamma_k \leq 0\}
\end{equation}
a simplified form of the initial positive sequence
estimator~\cite{geyer1992practical}.
\begin{table}[t]
  \centering
  \caption{Likelihood on lateral ($\pm5$\,mm) and tilt ($\pm3^\circ$)
    grids through the ground truth, tempered and summed over the first
    150 updates. Bias: distance of the likelihood-weighted mean from
    the ground truth; $\sigma$: weighted std. Values are pooled over model
  seeds and episodes.}
  \label{tab:loglik_eval}
  \small
  \setlength{\tabcolsep}{3pt}
  \begin{tabular}{lrrrr}
    \toprule
    & \multicolumn{2}{c}{Bias} & \multicolumn{2}{c}{$\sigma$} \\
    \cmidrule(lr){2-3} \cmidrule(lr){4-5}
    Likelihood & Lat. (mm) & Tilt ($^\circ$) & Lat. (mm) & Tilt ($^\circ$) \\
    \midrule
    \multicolumn{5}{l}{\emph{aleatoric + epistemic}} \\
    Motion & 0.45\,{\scriptsize$\pm$0.28} &
    0.11\,{\scriptsize$\pm$0.02} & 0.52\,{\scriptsize$\pm$0.07} &
    0.09\,{\scriptsize$\pm$0.03} \\
    Wrench & 0.59\,{\scriptsize$\pm$0.41} &
    0.17\,{\scriptsize$\pm$0.12} & 0.49\,{\scriptsize$\pm$0.14} &
    0.22\,{\scriptsize$\pm$0.02} \\
    Joint & 0.48\,{\scriptsize$\pm$0.20} &
    0.12\,{\scriptsize$\pm$0.02} & 0.30\,{\scriptsize$\pm$0.09} &
    0.09\,{\scriptsize$\pm$0.00} \\
    \midrule
    \multicolumn{5}{l}{\emph{aleatoric}} \\
    Motion & 0.45\,{\scriptsize$\pm$0.28} &
    0.11\,{\scriptsize$\pm$0.02} & 0.52\,{\scriptsize$\pm$0.07} &
    0.09\,{\scriptsize$\pm$0.03} \\
    Wrench & 0.57\,{\scriptsize$\pm$0.39} &
    0.15\,{\scriptsize$\pm$0.11} & 0.49\,{\scriptsize$\pm$0.13} &
    0.22\,{\scriptsize$\pm$0.02} \\
    Joint & 0.48\,{\scriptsize$\pm$0.20} &
    0.12\,{\scriptsize$\pm$0.03} & 0.31\,{\scriptsize$\pm$0.08} &
    0.09\,{\scriptsize$\pm$0.00} \\
    \midrule
    \multicolumn{5}{l}{\emph{unscaled aleatoric}} \\
    Motion & 0.58\,{\scriptsize$\pm$0.26} &
    0.16\,{\scriptsize$\pm$0.00} & 0.57\,{\scriptsize$\pm$0.04} &
    0.13\,{\scriptsize$\pm$0.00} \\
    Wrench & 0.65\,{\scriptsize$\pm$0.14} &
    0.16\,{\scriptsize$\pm$0.06} & 0.25\,{\scriptsize$\pm$0.14} &
    0.15\,{\scriptsize$\pm$0.08} \\
    Joint & 0.61\,{\scriptsize$\pm$0.13} &
    0.17\,{\scriptsize$\pm$0.10} & 0.23\,{\scriptsize$\pm$0.10} &
    0.09\,{\scriptsize$\pm$0.03} \\
    \midrule
    \multicolumn{5}{l}{\emph{no correlations}} \\
    Motion & 0.62\,{\scriptsize$\pm$0.35} &
    0.13\,{\scriptsize$\pm$0.01} & 0.66\,{\scriptsize$\pm$0.09} &
    0.09\,{\scriptsize$\pm$0.00} \\
    Wrench & 0.61\,{\scriptsize$\pm$0.53} &
    0.27\,{\scriptsize$\pm$0.03} & 0.57\,{\scriptsize$\pm$0.04} &
    0.22\,{\scriptsize$\pm$0.02} \\
    Joint & 0.57\,{\scriptsize$\pm$0.41} &
    0.11\,{\scriptsize$\pm$0.02} & 0.39\,{\scriptsize$\pm$0.14} &
    0.10\,{\scriptsize$\pm$0.02} \\
    \midrule
    \multicolumn{5}{l}{\emph{constant noise}} \\
    Motion & 0.68\,{\scriptsize$\pm$0.29} &
    0.13\,{\scriptsize$\pm$0.03} & 0.64\,{\scriptsize$\pm$0.07} &
    0.13\,{\scriptsize$\pm$0.01} \\
    Wrench & 0.56\,{\scriptsize$\pm$0.56} &
    0.25\,{\scriptsize$\pm$0.00} & 0.58\,{\scriptsize$\pm$0.09} &
    0.21\,{\scriptsize$\pm$0.01} \\
    Joint & 0.53\,{\scriptsize$\pm$0.40} &
    0.13\,{\scriptsize$\pm$0.01} & 0.37\,{\scriptsize$\pm$0.12} &
    0.13\,{\scriptsize$\pm$0.01} \\
    \bottomrule
  \end{tabular}
\end{table}

\balance
\bibliographystyle{IEEEtran}
\bibliography{refs}

\end{document}